\documentclass[11pt,a4paper]{article}
\usepackage[]{acl2020}
\usepackage{times}
\usepackage{booktabs}
\usepackage{latexsym}
\usepackage{multirow}

\usepackage{microtype}

\aclfinalcopy 

\usepackage{xspace}
\usepackage{graphicx}
\usepackage{xcolor}
\usepackage{multirow}
\usepackage{pgfplots}
\pgfplotsset{compat=1.14}

\newcommand\tencdec{\textsc{TranS2S}\xspace}
\newcommand\bertsum{\textsc{BertSum}\xspace}
\newcommand\bencdec{\textsc{BertS2S}\xspace}
\newcommand\rencdec{\textsc{RoBERTaS2S}\xspace}
\newcommand\quriousz{\textsc{QuriousZ}\xspace}
\newcommand\qurious{\textsc{Qurious}\xspace}

\newcommand\gold{\textsc{Gold}\xspace}

\definecolor{forestgreen}{HTML}{009B55}
\definecolor{sepia}{HTML}{671800}
\definecolor{midnightblue}{HTML}{006795}
\definecolor{orangered}{HTML}{ED135A}

\title{\qurious: Question Generation Pretraining for Text Generation}

\author{
  Shashi Narayan \\
  Google Research \\
  {\small \tt shashinarayan@google.com} \\ \And
  Gon\c{c}alo Simoes \\
  Google Research \\
  {\small \tt gsimoes@google.com} \\\And 
  Ji Ma \\
  Google Research \\
  {\small \tt maji@google.com} \\\AND
  Hannah Craighead \\
  Google \\
  {\small \tt craighead@google.com} \\\And 
  Ryan Mcdonald \\
  Google Research \\
  {\small \tt ryanmcd@google.com}
}

\date{}

\begin{document}
\maketitle

\begin{abstract}
Recent trends in natural language processing using pretraining have shifted focus towards pretraining and fine-tuning approaches for text generation. Often the focus has been on task-agnostic approaches that generalize the language modeling objective. We propose question generation as a pretraining method, which better aligns with the text generation objectives. Our text generation models pretrained with this method are better at understanding the essence of the input and are better language models for the target task. 
When evaluated on two text generation tasks, abstractive summarization and answer-focused question generation, our models result in state-of-the-art performances in terms of automatic metrics. Human evaluators also found our summaries and generated questions to be more natural, concise and informative.
\end{abstract}

\section{Introduction}

Unsupervised or semi-supervised pretrained encoder-decoder models are quickly becoming the standard for text generation  \cite{Khandelwal_2019,unilm_arxiv19,mass_icml19,berts2s,bart}, following the success of pretraining methods on popular Natural Language Understanding (NLU) benchmarks \cite{elmo,bert,gpt,gpt2,xlnet_arxiv19,roberta}.
For text generation, most models focus on task-agnostic pretraining tasks that generalize the language modelling objective by (i) combining the masked language model objective with left-to-right language modelling objective \cite{unilm_arxiv19} or (ii) reconstructing the corrupted input text using a sequence-to-sequence denoising autoencoder \cite{mass_icml19,bart}. These models have set new state-of-the-art results on a wide variety of text generation tasks such as summarization, sentence splitting and sentence fusion \cite{berts2s,bart}.

\begin{figure}[t!]
  \center{
  \setlength{\tabcolsep}{0.0in}
    \begin{tabular}{p{7.7cm}}
    \hline 
      
    \textbf{Answer Passage}: Mid March is still cold here in Europe. Some areas has also snow. So you need winter clothes.  But have fun, fun and fun. Europe is so nice. \\
    \textbf{Question}: What will be the temperature in Europe during March middle?
    \\ \hline
    \end{tabular}     
  }
  \caption{An example QA pair from Yahoo Answers.}\label{fig:question}
  \vspace{-0.6cm}
\end{figure}

In this paper, we investigate a pretraining objective that is better tied to challenges involved in text generation, specifically
understanding (i.e., identifying important content) and realization (i.e., generating the text).
We propose \qurious, a \textsc{QU}estion gene\textsc{R}ation pretra\textsc{I}ning \textsc{O}bjective which pretrains text generation models to generate questions conditioning on an answer passage or a document. Key advantages of our method are that (i) data for question generation can be easily crawled abundantly from community QA platforms such as Yahoo Answers, Quora and Stack Overflow, and more importantly, (ii) text generators trained to generate a question which can be answered from a document or a passage, will capture the salient terms or concepts expressed in the input, and will learn to aggregate and paraphrase from the input. 
Figure~\ref{fig:question} shows an example answer-question pair used for our pretraining reflecting on the latter point.

In this paper, we experiment with Transformer-based sequence-to-sequence models that are compatible with publicly available pretrained BERT~\cite{bert} and RoBERTa \cite{roberta} checkpoints, except these models were pretrained for question generation from an associated text. However, the question generation pretraining objective is model agnostic.
Improved pretraining objectives have been studied before, e.g., task agnostic \cite{xlnet_arxiv19,ALBERT}, multilingual \cite{Pires_2019} or domain targeted \cite{Lee_2019}. Perhaps the work closest to ours is 
\newcite{matchtheblanks} who study task-specific pretraining objectives for relation extraction. Additionally, \newcite{alberti-etal-2019-synthetic} use question generation to increase training data for QA. However, such task-specific objectives, and specifically question generation, have not been exploited for generation tasks.

Figure~\ref{fig:examples} demonstrates the benefit of our pretraining objective for summarization. \quriousz, a zero-shot variant of \qurious pretrained for question generation and without any supervision for summarization, generates questions for documents centered around their reference summaries; it appears that \quriousz simulates summarization experts in terms of selecting what source content is most relevant for a summary.
We hypothesize that the finetuning of \quriousz for summarization will guide models to focus on the salient content in the document and generate summaries that are concise and informative.
We can also observe that \qurious
generates summaries that are closer to reference summaries than those generated by \rencdec, which does not use pretraining for question generation.

\begin{figure}[t!]
  \center{
  \setlength{\tabcolsep}{0.0in}
    \begin{tabular}{p{7.7cm}}
    \hline 
      
    \textbf{\gold}: Former Beatle Sir Paul McCartney has topped the Sunday Times rich list of musicians with his $\pounds$730m fortune. \\
    
    \textbf{\rencdec}: Sir Paul McCartney has been named as Britain's richest man in the Sunday Times rich list. \\
    
    \textbf{\quriousz}: Who is the richest musician in the world? \\
    \textbf{\qurious}: Sir Paul McCartney has been named the richest man in the UK, with his wealth totalling $\pounds$730m, according to the Sunday Times rich list. \\ \hline
    
    \textbf{\gold}: Pope Francis will go to Africa for the first time this week, visiting a refugee camp, a slum and a mosque. \\
    
    \textbf{\rencdec}: Pope Francis has a big issue with the pope's decision to visit the Central African Republic in the middle of his first trip to the continent. \\
    
    \textbf{\quriousz}: What will Pope Francis talk about during his trip to Kenya? \\
    \textbf{\qurious}: Pope Francis will head to Kenya for his first visit to Africa since taking office in November. \\ \hline
  
    
    
  
    
    
    
    \end{tabular}     
  }
  \caption{
  Analysis of summarization models: the reference summary (\gold),
  a task-agnostic pretrained Seq2Seq model (\rencdec; \citeauthor{roberta,berts2s}, \citeyear{berts2s}) and one pretrained with a question generation objective (\qurious and a zero-shot variant \quriousz).
  }\label{fig:examples}
  \vspace{-0.6cm}
\end{figure}

The main contributions of this work are four-fold. First, we propose question-generation as a pretraining objective for text generation. Second, we demonstrate the effectiveness of our method on abstractive summarization by achieving a new state-of-the-art result on the extreme summarization task \cite{narayan-etal-2018-xsum}. Third, we experiment with answer-focused question generation task focusing on two datasets, SQuAD \cite{squad} and Natural Questions \cite{NQ}, and demonstrate that our pretrained model generates questions that are more natural and informative, in terms of both automatic and human evaluations. 
Finally, we empirically demonstrate that the reciprocity of 
question generation as a pretraining objective to text generation tasks makes our models robust to low-resource scenarios.

\section{Question Generation Pretraining}

\qurious is designed for sequence-to-sequence models and aims to learn improved representations for text generation, which requires both understanding and realization, as opposed to task-agnostic pretraining objectives \cite{bert}.  


\paragraph{Data for Pretraining.} 
In this work, we collect 2 million English question-answer pairs from community question-answering resources such as StackExchange (53.8\% of total, 175 subdomains), Yahoo! Answers (45.9\% of total, 24 subdomains) and Zhidao Baidu (0.3\%). These forums have been widely used before in community question answering \cite{commqacikm,nakov-etal-2017-semeval,commqaieee}. In particular, we follow \newcite{commqacikm} to mine data from community QA websites. Main differences are that \newcite{commqacikm} mine data from two community QA websites and train answer passage selection models, whereas, we (1) use a different set of websites and (2) use it for question generation pretraining. 
To ensure the quality of posts we only select English answer-question pairs that were positively rated by at least one user. Finally, the average lengths of questions and answers in our dataset are 11.64 tokens and 155.44 tokens, respectively.

A major advantage of \qurious is that large amounts of pretraining data can be obtained for free, and annotations grow as long as people ask/answer questions on the internet. 
Moreover, real information-seeking questions are typically condense and natural, thus better suited for summarization than datasets such as SQuAD \cite{squad}, where questions are not naturally occurring and contain high lexical and syntactic overlap with the answer passage. 

\paragraph{Pretraining Text Generation Models.}
We apply \qurious to a sequence-to-sequence architecture where both encoder and decoder are composed of Transformer layers \cite{transformer}.
We have experimented with base and large versions of the Transformer layer; the base model has both encoder and decoder with 12 layers, a hidden size of 768, filter size of 3072, and 12 attention heads, whereas, the large model, with 24 layers, a hidden size of 1024, filter size of 4096, and 16 attention heads. 
During pretraining, the input answers were truncated to 512 tokens and the length of the questions was limited to 64 tokens. 
We also allow our encoder and decoder to warm-start the Transformer layer using public BERT \cite{bert} and its variant RoBERTa \cite{roberta} checkpoints. 
Following \newcite{berts2s}, we share the parameters between encoder and decoder for all our models. We used a global batch size of 128 document-summary pairs with the standard cross entropy loss.



\paragraph{Fine-tuning Text Generation Models.}

We fine tune our model for two text generation tasks: abstractive document summarization and answer-focused question generation. For abstractive summarization, the encoder takes a document as input and generates its summary as output. 
For answer-focused question generation, earlier work \cite{duan-etal-2017-question,subramanian-etal-2018-neural,nema-etal-2019-lets} has mostly focused on the factoid-based question answering dataset such as SQuAD \cite{rajpurkar-etal-2016-squad}. Unlike our question generation pretraining, the answer passage here can be open-ended and not necessarily a direct response to a question.
We follow \newcite{nema-etal-2019-lets} and use the target answer span together with the passage (with a separator between them) as input to generate a specific question.




\section{Experiments and Results}

\subsection{Abstractive Document Summarization}
\label{sec:abstractive_sum}

We evaluate our model on the BBC extreme summarization (XSum; \citeauthor{narayan-etal-2018-xsum},  \citeyear{narayan-etal-2018-xsum}). Documents in this dataset are accompanied by their single-sentence summaries with a high level of abstractiveness, and generating them requires document-level inference, abstraction and paraphrasing. The dataset consists of 204k, 11k and 11k document-summary training, validation and test pairs. 

\begin{table}[t!]
\centering
\footnotesize
\begin{tabular}{l|ccc}
\hline
\textbf{Models} & \textsc{r1} & \textsc{r2} & \textsc{rL}\\ \hline
\tencdec (12) & 30.90 & 10.23 & 24.24 \\ 
\bertsum (12) & 38.81 & 16.50 & 31.27 \\
\bencdec (12) &  38.52 & 16.12 & 31.13 \\ 
\bencdec (24) & 38.93 & 16.35 & 31.52 \\ 
\rencdec (12) & 39.87 & 17.50 & 32.37 \\
\rencdec (24) & 41.45 & 18.79 & 33.90 \\ \hline 
\multicolumn{4}{c}{Our models} \\ \hline
\qurious (-\textsc{RoBERTa}, 12) & 32.31 & 11.69 & 25.63 \\
\qurious (-\textsc{RoBERTa}, 24) & 31.96 & 11.60 & 25.47\\
\qurious (12) & 40.50 & 18.21 & 32.94 \\
\qurious (24) & \textbf{42.53} & \textbf{19.88} & \textbf{34.93} \\
\hline
\quriousz (-\textsc{RoBERTa}, 24) & 10.01 & 1.07 & 8.54 \\
\quriousz (24) & 16.76 & 3.67 & 13.95 \\ \hline
\end{tabular}
\caption{
ROUGE F$_1$ scores for extreme summarization. The models in the top block are not pretrained for question generation. See text for discussion. \textsc{r1/2/L} is \textsc{rouge1/2/L}.}
\label{tab:summarization}
\vspace{-0.6cm}
\end{table}

\paragraph{Automatic Evaluation}

We report on the ROUGE $F_1$ scores \cite{rouge} in Table~\ref{tab:summarization}. Our main baseline is a transformer-based Seq2Seq model, \tencdec, initialized with a public BERT \cite{bert} checkpoint (\bencdec) as reported in \citet{berts2s}. We also report numbers for a second BERT-based transformer model, \bertsum \cite{liu-lapata-2019-text}.
Finally, we experimented with using a RoBERTa \cite{roberta} checkpoint. This model, \rencdec, significantly improves over the state-of-the-art \bencdec and \bertsum. 

Following the advantages of \rencdec over \bencdec, \qurious initializes with the RoBERTa checkpoint and pretrains with the question generation objective, before fine tuning for extreme summarization. We also perform an ablation study where we do not initialize our model with the RoBERTa checkpoint (\qurious -\textsc{RoBERTa}).  \quriousz is not fine tuned for extreme summarization, it behaves as a question generation model which takes a document as input and generate a question. It assesses how close the generated questions get to the reference summary.

As can be seen in Table~\ref{tab:summarization}, the question generation pretraining in \qurious (-\textsc{RoBERTa}) improves over \tencdec across all ROUGE scores (improvement of 1.42 points on average). \qurious with the RoBERTa initialization outperforms its counterpart \rencdec; this improvement is consistent for both base (12) and large (24) models. \qurious (24) achieves a new state-of-the-art on extreme summarization outperforming earlier  model \bencdec (24; \citeauthor{berts2s}, \citeyear{berts2s}) by 3.51 average ROUGE points. Interestingly, the question generation pretraining elevates the performance of RoBERTa initialized Seq2Seq model for summarization; our pretraining objective should also supplement recent pretraining schemes \cite{unilm_arxiv19,mass_icml19,bart} for summarization.



\begin{table}[t!]
\centering
\footnotesize
\begin{tabular}{l|ccccc}
\hline
\textbf{Models} & \textsc{b1} & \textsc{b2} & \textsc{b3} & \textsc{b4} & \textsc{rl}\\ \hline
\multicolumn{6}{c}{SQuAD} \\ \hline
\citeauthor{zhao-etal-2018-paragraph} & 45.1 & 29.6 & 21.6 & 16.4 & 44.5 \\
\citeauthor{nema-etal-2019-lets} & 46.4 & 30.7 & 22.4 & 17.0 & 45.0 \\
\rencdec  & 45.6 & 29.4 & 20.7 & 15.1 & 45.0 \\
\qurious  & \textbf{47.4} & \textbf{32.0} & \textbf{23.5} & \textbf{17.8} & \textbf{46.7} \\ \hline
\multicolumn{6}{c}{NQ} \\ \hline
\rencdec  & 55.4 & 42.8 & 34.0 & 27.1 & 53.5 \\
\qurious & \textbf{57.4} & \textbf{45.1} & \textbf{36.3} & \textbf{29.3} & \textbf{55.4} \\ \hline
\end{tabular}
\caption{
Question generation results on SQuAD and Natural Questions (NQ) datasets. For each model, we choose its best performing variant from Table~\ref{tab:summarization} for this task. \textsc{b1-4} is \textsc{bleu1-4}; \textsc{rl} is \textsc{rouge-l}.}
\label{tab:qgen}
\vspace{-0.6cm}
\end{table}

\subsection{Answer-focused Question Generation}
\label{answer_focused_qg}

For the Question Generation task, we evaluate our models on two factoid-based question answering datasets: SQuAD~\cite{rajpurkar-etal-2016-squad} and Natural Questions (NQ; \citeauthor{kwiatkowski-etal-2019-natural}, \citeyear{kwiatkowski-etal-2019-natural}). 
For SQuAD, we use the whole paragraph as input passage and not just the sentence containing the answer as it often requires the whole paragraph as context in order to generate high quality questions. In total, this dataset is composed of 87K training examples and 10K development examples.
For NQ, we use the provided long answer as input passage. 
We only keep those that are paragraphs and filter out list and table based long answers.  
We further filter Yes/No questions and also questions that are not answerable, this results in a training set of 95K examples and a development set 3.6K examples. 
To the best of our knowledge, we are the first to use the NQ dataset for the question generation task.\footnote{\newcite{alberti-etal-2019-synthetic}  used the NQ dataset to construct  synthetic question-answer corpora to train QA models.}


\paragraph{Automatic Evaluation.}

We choose our best performing model from Table~\ref{tab:summarization} for this task.
We report on the BLEU \cite{papineni-etal-2002-bleu} and ROUGE-L $F_1$ \cite{rouge} scores, and results are listed in Table~\ref{tab:qgen}.
Table~\ref{tab:qgen} exhibits a similar pattern as that in Table~\ref{tab:summarization}: \qurious consistently improve model performance over \rencdec on both SQuAD and NQ.  For SQuAD, \qurious also achieves substantial improvement over previous state of the art \cite{nema-etal-2019-lets}.  

Since both NQ and our pre-training dataset consist of real information-seeking questions, we expect \qurious to perform better on NQ than SQuAD.
It turns out that all models trained on NQ achieve much higher scores than their SQuAD counterparts.
This result suggests that naturalness of the question is a more important factor in affecting system performance.  






\subsection{Human Evaluations}

In addition to automatic evaluation using ROUGE and BLEU, we also evaluated system output by eliciting human judgments for both summarization and question generation. The study was conducted on the Amazon Mechanical Turk platform using Best-Worst Scaling, a less labor-intensive alternative to paired comparisons \cite{louviere1991best,louviere2015best}. For summarization, participants were presented with a document and summaries generated from two out of five systems and were asked to decide which summary was better than the other in order of informativeness (does the summary capture important information in the document?) and fluency (is the summary written in well-formed English?). For question generation, participants were presented with an answer passage, a factoid answer and questions generated from two systems and were asked to decide which question is more (i) natural (is the summary fluent and written in well-formed English?) and (ii) correct (is the question correct for the factoid answer given the passage?). In all cases, we allowed ties when both predictions were the same. Additionally, for correctness, we allowed a tie when both questions were equally correct or incorrect. We randomly selected 30 documents from the XSum test set for summarization and 30 answer-passage pairs each for question generation from SQuAD and from Natural Questions. We collected judgments from three different participants for each comparison. The order of summaries were randomized per document and the order of documents per participant. The score of a system was computed as the percentage of times it was chosen as best minus the percentage of times it was selected as worst. The scores range from -1 (worst) to 1 (best). Some of the sample predictions used in human evaluations are presented in the appendix.



\begin{table}[t!]
\centering
\footnotesize
\begin{tabular}{l|c|c c | c c}
\hline
\multirow{3}{*}{\textbf{Models}} & \multirow{2}{*}{\textbf{XSum}} & \multicolumn{4}{|c}{QGen} \\ 
 &  & \multicolumn{2}{c|}{SQuAD} & \multicolumn{2}{c}{NQ} \\ 
&  quality &  nat. & corr. & nat. & corr. \\ \hline
\citeauthor{nema-etal-2019-lets} & -- & -0.23 & -0.24 & -- & -- \\
\rencdec & -0.25 & 0.01 & -0.03 & -0.15 & -0.13 \\
\qurious & \textbf{0.14} & \textbf{0.12} & \textbf{0.19} & \textbf{0.11} & 0.02 \\
\gold & 0.11 & 0.10& 0.08 & 0.04 & \textbf{0.17} \\ \hline
\end{tabular}
\caption{
Human evaluation results for summarization assessing summary `quality'  and answer-focused question generation assessing naturalness (nat.) and correctness (corr.) of questions.}
\label{tab:human}
\vspace{-0.3cm}
\end{table}

\qurious outperformed \rencdec across all tasks. Interestingly, it even performed better than human-authored summaries or questions with a single exception of the correctness assessment of questions on the NQ dataset.  


We carried out pairwise comparisons between all models to assess whether system differences are statistically significant (using a one-way ANOVA with posthoc Tukey HSD tests; $p < 0.01$). 
For summarization, \rencdec is significantly different from both \qurious and \gold. For SQuAD, \citeauthor{nema-etal-2019-lets} is significantly different from all other systems on naturalness and correctness, and \rencdec is significantly different from \qurious on correctness. For NQ, \rencdec is significantly different from 
\qurious on naturalness and from \gold on correctness. All other differences are not statistically significant.

\begin{figure}[t!]
  \center{
  \setlength{\tabcolsep}{0.0in}
    \begin{tabular}{p{7.7cm}}
    \hline 
    \textbf{SQuAD Passage}: The acme of the horizontal engine was the Corliss steam engine, patented in 1849 ... \\
    \textbf{Answer} 1849 \\
    \textbf{\qurious}: When was the Corliss steam engine patented? (Prefered) \\
    \textbf{\gold}: In what year was the Corliss engine patented?
    \\ \hline
    \textbf{NQ Passage}: The Last Supper (...) is a late 15th-century mural painting by Leonardo da Vinci ... \\
    \textbf{Answer} Leonardo da Vinci \\
    \textbf{\qurious}: Who painted The Last Supper in the Louvre? \\
    \textbf{\gold}: Who painted the world famous painting The Last Supper? (Prefered)
    \\ \hline
    \end{tabular}     
  }
  \vspace{-0.1in}
  \caption{\qurious predictions on the SQuAD and NQ datasets.}\label{fig:qgenex}
  \vspace{-0.6cm}
\end{figure}

The difference in performance of \qurious on SQuAD and NQ stem from how these datasets are created. For SQuAD, human annotators started with the passage and wrote the questions, many times resorting to paraphrasing with copying involved. For NQ, the dataset creation process started with the questions, making them more natural and harder for a model to learn by copying. Consequently, \qurious appears to hallucinate more when fine tuned on NQ than when fine tuned on SQuAD (see examples in Figure~\ref{fig:qgenex}). Regardless, the differences between \qurious and \gold are not statistically significant. 

\subsection{Sample Efficiency Experiments}

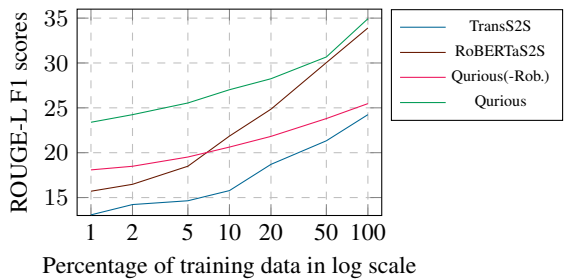
\begin{figure}[t!]
  \footnotesize
  \center{
    \begin{tikzpicture}
      \begin{axis}[
        width=2.2in,
        height=1.7in,
        title={},
        xlabel={Percentage of training data in log scale},
        ylabel={ROUGE-L F1 scores},
        xmin=-1, xmax=21,
        ymin=13, ymax=36,
        xtick={0, 3, 7, 10, 13, 17, 20},
        ytick={10, 15, 20, 25, 30, 35},
        xticklabels={1, 2, 5, 10, 20, 50, 100},
        legend pos=outer north east,
        xmajorgrids=true,
        ymajorgrids=true,
        grid style=dashed,
        legend style={font=\tiny},
        ]
        \addplot [midnightblue] table[x=TrainingData,y=TransS2S]{\rougel};
        \addplot [sepia] table[x=TrainingData,y=RoBERTaS2S]{\rougel};
        \addplot [orangered] table[x=TrainingData,y=QuirousNoRoBERTa]{\rougel};
        \addplot [forestgreen] table[x=TrainingData,y=Qurious]{\rougel};
        \legend{TransS2S, RoBERTaS2S, Qurious(-Rob.),Qurious}
      \end{axis}
    \end{tikzpicture}
  }
  \caption{
  Question generation pretraining is sample efficient for summarization.\label{fig:sample_eff}}
  \vspace{-0.6cm}
\end{figure}

Finally, we evaluated how \qurious performs in low-resource scenarios by performing sample efficiency experiments. We focus on the extreme summarization task (\S\ref{sec:abstractive_sum}), but test each model when trained using only a subset
of the supervised fine-tuning data. Figure~\ref{fig:sample_eff} presents our results. What is interesting is that both \qurious and \qurious(-\textsc{RoBERTa}) outperform \rencdec at the very low resource settings, suggesting that content selection driven pretraining objectives are more important than task-agnostic masking objectives. Even more interesting is that the \qurious model significantly outperforms \rencdec until about 50\% of the training data is consumed, at which point performance converges, though with \qurious on top. This suggests that the optimal configuration is pretraining objectives that include both content selection (question generation) and knowledge-based (language-model) criteria. Intuitively this makes sense, a good summary should have rich knowledge of content in order to know how to select content and realize it accurately.

\pgfplotstableread[row sep=\\,col sep=&]{
  TrainingData & TransS2S & RoBERTaS2S & QuirousNoRoBERTa & Qurious\\ 
  0 & 13.071 & 15.706 & 18.084 & 23.392 \\
  3 & 14.22 & 16.477 & 18.488 & 24.239 \\
  7 & 14.643 & 18.493 & 19.518 & 25.542 \\
  10 & 15.775 & 21.85 & 20.626 & 27.017 \\
  13 & 18.7 & 24.85 & 21.818 & 28.243 \\
  17 & 21.327 & 30.037 & 23.805 & 30.669\\
  20 & 24.24 & 33.90 & 25.47 & 34.93 \\
}\rougel

\section{Conclusion}

In this paper, we proposed a question generation pretraining objective for text generation. When evaluated for summarization and answer-focused question generation tasks, our model generated summaries and questions, respectively, that were more natural and informative in terms of automatic and human evaluations. In the future, we would like to explore if the question generation pretraining objective can be beneficial for other text generation and language understanding tasks.


\bibliography{question_generation}
\bibliographystyle{acl_natbib}

\appendix

\section{Summarization Outputs}
\label{sec:app-sum}

Figure~\ref{fig:sum_examples} shows examples of BBC articles and their extreme summaries.

\begin{figure*}[t!]
  \center{
    \begin{tabular}{p{15cm}}
    \hline 
      
    \textbf{\gold}: Former Beatle Sir Paul McCartney has topped the Sunday Times rich list of musicians with his $\pounds$730m fortune. \\ \hline
    
    \textbf{Document}:  Sir Paul is worth an estimated $\pounds$20m more than last year and enjoys a significant boost from his American heiress wife's $\pounds$150m stake in her family's US trucking business. It puts him well ahead of his nearest rival on the list, Andrew Lloyd Webber, who is estimated to be worth $\pounds$650m. The full list will be published by the newspaper on 26 April. Of the 1,000 richest people in the UK and the 250 wealthiest in Ireland, the list puts Irish band U2 at third place with $\pounds$431m. Pop veteran Sir Elton John and Rolling Stones 'frontman Sir Mick Jagger follow with their fortunes, thought to be worth $\pounds$270m and $\pounds$225m respectively. 1. Sir Paul McCartney and Nancy Shevell $\pounds$730m (Rest of the article is abbreviated ...) \\ \hline
    
    \textbf{\tencdec}: One of the richest people in the UK has topped the list of the richest people in the world. \\
    
    \textbf{\qurious(-\textsc{RoBERTa})}: The Rolling Stones have been named the richest young band in the UK this year. \\
    
    \textbf{\rencdec}: Sir Paul McCartney has been named as Britain's richest man in the Sunday Times rich list. \\
    \textbf{\qurious}: Sir Paul McCartney has been named the richest man in the UK, with his wealth totalling $\pounds$730m, according to the Sunday Times rich list. \\ \hline  
    \vspace{0.1cm} \\
    \hline

    \textbf{\gold}: Islanders on Skye have demanded greater availability of public toilets after complaints some visitors to the Isle are relieving themselves outside. \\ \hline
    
    \textbf{Document}: There have been incidents reported at scenic spots where public conveniences are lacking or have been closed down. In Uig, where many of the complaints have been raised, the local authority-run toilets have been out of order since the beginning of the year. Highland Council said it was seeking quotes for the repair work needed. The availability of toilets on Skye has been raised previously. In 2011, Highland Council received complaints about people urinating and defecating outdoors at Staffin where public toilets were closed as part of cost cutting. \\
    
    \textbf{\tencdec}: A council has asked people not to keep their toilets in a bid to save money. \\
    
    \textbf{\qurious(-\textsc{RoBERTa})}: Highland Council is calling on public complaints about a possible route for people to urinating on Skye. \\
    
    \textbf{\rencdec}: Highland Council has commissioned a review of public toilets and public toilets on Skye. \\
    
    \textbf{\qurious}: Highland council is seeking information about problems with public toilets on Skye. \\ \hline

    \end{tabular}     
  }
  \caption{Example documents and summarization model predictions.}\label{fig:sum_examples}
\end{figure*}

\section{Squad Outputs}
\label{sec:app-squad}

Figure~\ref{fig:examples_squad_human_eval} shows examples of SQuAD input passages, answer spans and questions generated from them.

\begin{figure*}[t!]
 \center{
    \begin{tabular}{p{15cm}}
    \hline
    
    \textbf{Passage:} Under the terms of the Scotland Act 1978, an elected assembly would be set up in Edinburgh provided that the majority of the Scottish electorate voted for it in a referendum to be held on 1 March 1979 that represented at least 40\% of the total electorate. The 1979 Scottish devolution referendum to establish a devolved Scottish Assembly failed. (...)\\
    \textbf{Answer:} failed\\\\

    \textbf{\citeauthor{nema-etal-2019-lets}}: What happened to the 1979 Scottish devolution referendum in 1979? \\
      
    \textbf{\tencdec}: What percentage of the vote of Ireland was interpreted as a result of voting? \\
    
    \textbf{\rencdec}: How did the 1979 Scottish devolution referendum fail? \\
    
    \textbf{\qurious(-\textsc{RoBERTa})}: What did the Scottish assembly of Edinburgh vote to pass in 1979? \\
    
    \textbf{\qurious}: What was the result of the 1979 Scottish devolution referendum? \\
      
    \textbf{\gold}: How did trying to establish a devolved Scottish assembly go in 1979? \\ \hline \\ \hline
    
    \textbf{Passage:} Although lacking historical connections to the Middle East, Japan was the country most dependent on Arab oil. 71\% of its imported oil came from the Middle East in 1970. (...)\\
    
    \textbf{Answer:} 71\% \\\\

    \textbf{RefNet}: What percentage of its imported oil came from Japan? \\
      
    \textbf{\tencdec}: When did Japan make a national influence? \\
    
    \textbf{\rencdec}: How much oil did Japan's oil from the Middle East come in in 1970? \\
    
    \textbf{\qurious(-\textsc{RoBERTa})}: How much of the Middle East's oil was imported in Japan by the Middle East? \\
    
    \textbf{\qurious}: How much of Japan's imported oil came from the Middle East? \\
      
    \textbf{\gold}: How much imported oil came from the Middle East? \\\hline
  
    \end{tabular}     

}
 \caption{Examples produced by the answer-focused question Generation models on Squad. }\label{fig:examples_squad_human_eval}
\end{figure*}

\end{document}